%% file: adaptsign.tex
\documentclass[10pt,twocolumn,letterpaper]{article}

\usepackage{cvpr}              
\usepackage{amsmath}
\usepackage{amssymb}
\usepackage{multirow}
\usepackage{pifont}
\usepackage{makecell}
\usepackage{bbding}
\input{preamble}

\definecolor{cvprblue}{rgb}{0.21,0.49,0.74}
\usepackage[pagebackref,breaklinks,colorlinks,citecolor=cvprblue]{hyperref}

\def\paperID{782} 
\def\confName{CVPR}
\def\confYear{2024}

\title{Improving Continuous Sign Language Recognition with Adapted Image Models}

\author{Lianyu Hu, Tongkai Shi, Liqing Gao, Zekang Liu, Wei Feng\textsuperscript{\Envelope}
\and
College of Intelligence and Computing, Tianjin University, Tianjin 300350, China
\and
Code : \url{https://github.com/hulianyuyy/AdaptSign}
}
\begin{document}
\maketitle
\begin{abstract}
    The increase of web-scale weakly labelled image-text pairs have greatly facilitated the development of large-scale vision-language models (e.g., CLIP), which have shown impressive generalization performance over a series of downstream tasks. However, the massive model size and scarcity of available data limit their applications to fine-tune the whole model in downstream tasks. Besides, fully fine-tuning the model easily forgets the generic essential knowledge acquired in the pretraining stage and overfits the downstream data. To enable high efficiency when adapting these large vision-language models (e.g., CLIP) to performing continuous sign language recognition (CSLR) while preserving their generalizability, we propose a novel strategy (AdaptSign). Especially, CLIP is adopted as the visual backbone to extract frame-wise features whose parameters are fixed, and a set of learnable modules are introduced to model spatial sign variations or capture temporal sign movements. The introduced additional modules are quite lightweight, only owning 3.2\% extra computations with high efficiency. The generic knowledge acquired in the pretraining stage is well-preserved in the frozen CLIP backbone in this process. Extensive experiments show that despite being efficient, AdaptSign is able to demonstrate superior performance across a series of CSLR benchmarks including PHOENIX14, PHOENIX14-T, CSL-Daily and CSL compared to existing methods. Visualizations show that AdaptSign could learn to dynamically pay major attention to the informative spatial regions and cross-frame trajectories in sign videos. 
    \end{abstract}

\section{Introduction}
Sign language is one of the most commonly-used communication tools for the deaf community in their daily life. However, mastering this language is rather difficult and time-consuming for the hearing people, thus hindering direct communications between two groups. To relieve this problem, continuous sign language recognition (CSLR) progresses by sequentially translating image streams into a series of glosses\footnote{Gloss is the atomic lexical unit to annotate sign languages.} to express a complete sentence, more prospective towards bridging the communication gap.

Recently, the availability of large-scale web image-text pairs has greatly accelerated the development of vision-language models such as CLIP~\cite{radford2021learning}. These methods typically model multi-modal information in a contrastive way, by closing the distance of positive image-text pairs and pushing away others'~\cite{radford2021learning,alayrac2022flamingo,li2022blip,li2023blip,kim2021vilt}. Powered by the supervision of natural language, these models have shown impressive generalization performance over a series of downstream tasks~\cite{gao2021clip,fang2021clip2video,cheng2021improving,dzabraev2021mdmmt,YuWVYSW22} in a zero-shot manner.

Despite the excellent performance over novel concepts achieved by these models, it's still challenging to directly adapt them to downstream tasks for further performance boost following the traditional fine-tuning manner: (1) The massive model scale and scarcity of training data limit their applications to downstream scenarios, where the restricted computing resources and absence of data may not support fine-tuning the whole model. The situation may be more severe in video-related tasks due to the increase of computational costs and the challenge to collect labelled video data~\cite{lin2022frozen,jia2022visual}. (2) fully fine-tuning the model easily forgets the generic knowledge acquired in the pretraining stage and overfits the downstream data. The acquired well-generalized knowledge of the pretrained model is easily destroyed by fine-tuning on inadequate data which hurts the generalizability~\cite{ju2022robust}. 

\begin{figure}[t]
   \centering
   \includegraphics[width=\linewidth]{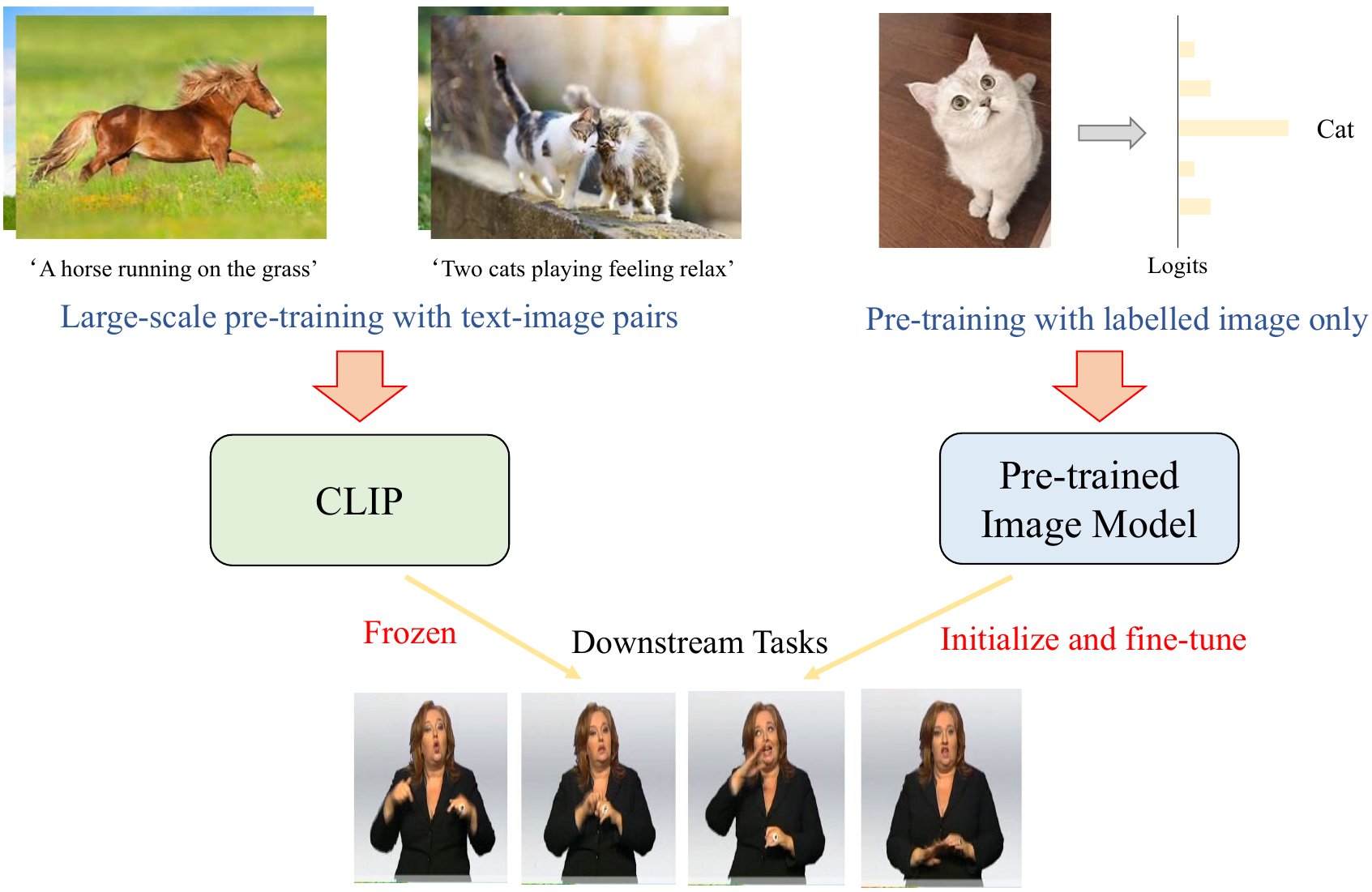}
   \caption{Illustration of the difference between our training pipeline and the commonly-used pretrain-and-finetune paradigm. }
   \label{fig1}
   \vspace{-5px}
\end{figure}

The above issues motivate us to find an efficient strategy to adapt these strong vision-language models to learning good video representations in continuous sign language recognition (CSLR), where data scarcity and limited computing resources are both encountered. To tackle the above problems, we propose a novel strategy (AdaptSign). Especially, we adopt a frozen CLIP model as the visual backbone, and stack several modules on top to learn more discriminative spatial sign features or model temporal sign correlations. The incurred extra computations by these modules are quite lightweight (3.2\%) compared to the frozen CLIP model, which enable high efficiency in this procedure. Specifically, We instantiate a new frame-level token as a query, with features from each CLIP block as keys and values to aggregate multiscale features. Adapters are appended in parallel within each CLIP block to update the intermediate spatial features in a residual way. We inject prefix embeddings before the visual features as learnable embeddings to model specific domain knowledge. To capture the temporal sign information, a  correlation module is introduced on top to model the cross-frame trajectories. Fig.~\ref{fig1} depicts the differences between our framework and the traditional fine-tuning pipeline. Extensive experiments show that despite being efficient, AdaptSign outperforms existing methods by a large margin across a series of CSLR benchmarks including PHOENIX14, PHOENIX14-T, CSL-Daily and CSL. Plentiful visualizations demonstrate AdaptSign is able to pay major attention to the informative spatial regions and cross-frame trajectories in sign videos.

\section{Related Work}
\subsection{Continuous Sign Language Recognition}
Sign language recognition methods can be roughly divided into isolated sign language recognition~\cite{tunga2021pose,hu2021signbert,hu2021hand} and continuous sign language recognition~\cite{pu2019iterative,cheng2020fully,cui2019deep,niu2020stochastic,Min_2021_ICCV} (CSLR), and we focus on the latter in this paper. CSLR aims to translate input images into corresponding glosses in a weakly-supervised way: only sentence-level label is provided. Earlier methods~\cite{gao2004chinese,freeman1995orientation} in CSLR always employ hand-crafted features or HMM-based systems~\cite{koller2016deepsign,han2009modelling,koller2017re,koller2015continuous} to perform temporal modelling and translate sentences step by step. Recently, the CTC loss~\cite{graves2006connectionist} is broadly used in recent CSLR methods~\cite{pu2019iterative,pu2020boosting,cheng2020fully,cui2019deep,niu2020stochastic,Min_2021_ICCV} to train deep networks in an end-to-end manner by sequentially aligning target sentences with input frames. These CTC-based methods first rely on a spatial extractor, which is often instantiated as a 2D CNN to extract frame-wise features, and then adopt a sequence model consisting of a 1D CNN and a LSTM for capturing temporal dependencies. However, several methods~\cite{pu2019iterative,cui2019deep} found in such conditions the spatial extractor is not well-trained and then present an iterative training strategy to relieve this problem, but consume much more computations. Some recent studies try to directly enhance the spatial extractor with visual supervision~\cite{Min_2021_ICCV,cheng2020fully, hao2021self}, squeezing more beneficial temporal features~\cite{hu2022temporal} or emphasize critical spatial features~\cite{hu2023self}. Adapting large vision-language models for CSLR faces two major problems, scarcity of available data (e.g., only 20k videos for commonly-used PHOENIX14~\cite{koller2015continuous} and CSL-Daily~\cite{zhou2021improving}) and huge computational costs during fine-tuning. We present a novel strategy to adapt these high-quality visual features for sign language understanding with high efficiency while preserving their generalizability. 

\subsection{Vision-Language Models} 
Powered by the abundance of image-text pairs collected from the web, large-scale vision-language methods have developed fast over the past several years. Earlier methods mainly rely on grid features~\cite{jiang2020defense,nguyen2020movie} or region proposals`~\cite{anderson2018bottom}' to align image features with text embeddings. In contrast to these methods, contrastive vision-language methods such as CLIP~\cite{radford2021learning} are trained by maximizing the feature similarity between positive image-text pairs, which learn more powerful visual features with aligned language semantics. Another advantage of CLIP is its impressive feature transferability, which shows promising results on a series of downstream visual tasks in a zero-shot manner. However, fine-tuning the whole model is still infeasible in some cases over the downstream tasks due to the scarcity of training data and incurred computations. In this paper, we design an efficient strategy to adapt these generic features to helping understand sign videos.

\subsection{Efficient Transfer Learning}
This set of work tries to efficiently transfer high-quality representations from pretrained models into downstream tasks, which are first explored in natural language processing and image recognition. Adapter series~\cite{houlsby2019parameter,hu2021lora,he2021towards} keep the pretrained model fixed and design adapters consisting of an MLP with residual connections to adjust output features. Some works explore prompt tuning~\cite{li2021prefix,lester2021power}, which append a learnable prompt before the input or the intermediate features to adapt the output features to specific tasks. In terms of the visual domain, CLIP-Adapter~\cite{gao2021clip} tries to calibrate the classifier on top of a frozen CLIP model via a residual way. VL-Adapter~\cite{sung2022vl} explores the setup of adapters in the multitask setting. Tip-Adapter~\cite{zhang2021tip} proposes a training-free adapter to align texts with images. VPT~\cite{jia2022visual} tests the choices of a set of visual prompts in visual tasks. EVL~\cite{lin2022frozen} designs a transformer decoder to learn more powerful spatial-temporal representations. Some works~\cite{cheng2021improving,fang2021clip2video} transfer the aligned image and text features for cross-modality tasks with a CLIP backbone. In contrast to previous methods that mostly focus on image understanding, we try to handle the data scarcity and incurred computations in video scenarios like CSLR.

\section{Method}
\begin{figure}[t]
   \centering
   \includegraphics[width=\linewidth]{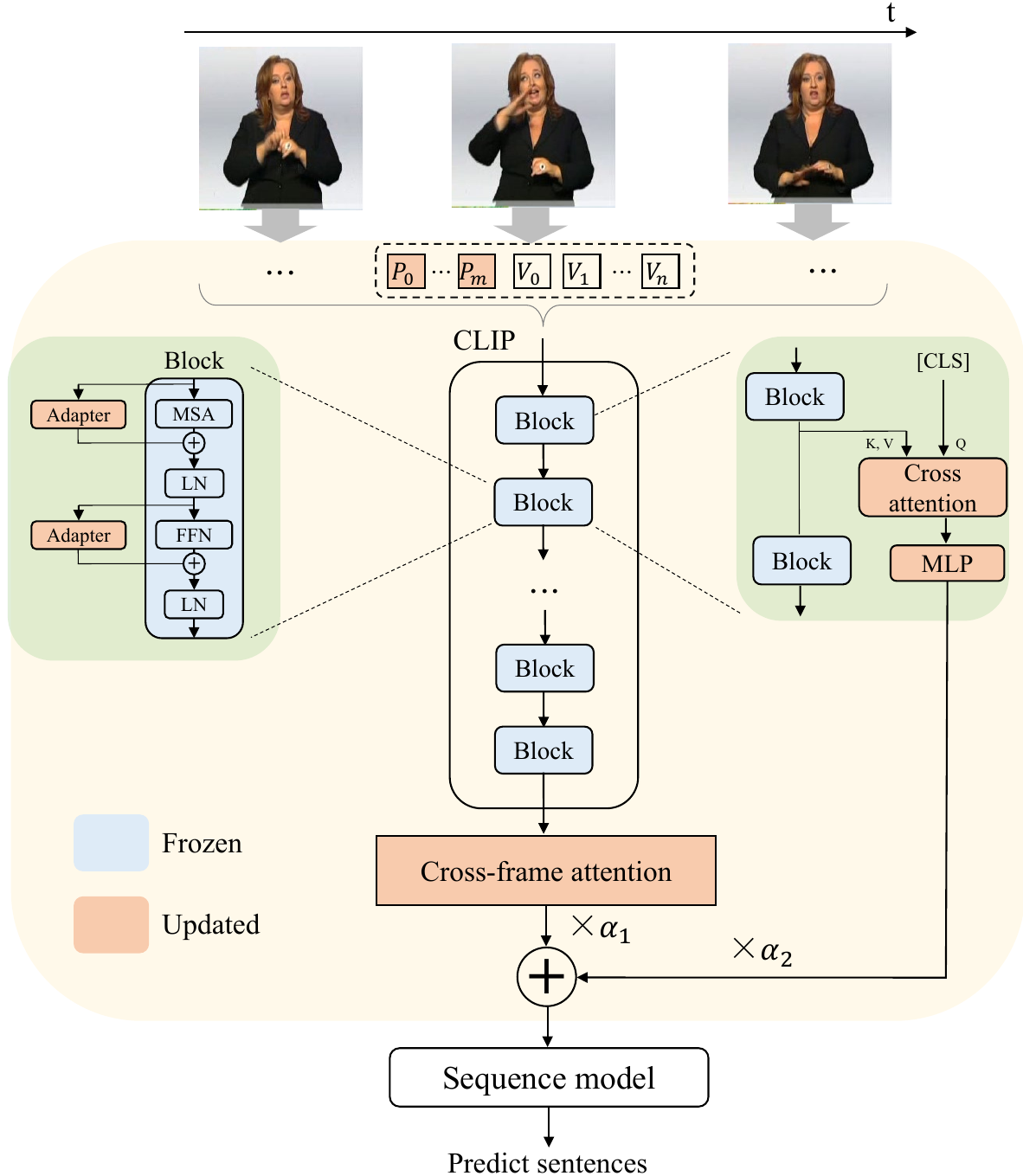}
   \caption{An overview for the framework of AdaptSign. We instantiate a frozen CLIP model as the backbone of spatial extractor, with several proposed modules on top for feature calibration. A sequence model is attached to perform sentence prediction following previous CSLR methods~\cite{Min_2021_ICCV,cheng2020fully, hao2021self,hu2022temporal,hu2023self}. The high-quality features of the frozen CLIP backbone are efficiently transferred through the proposed lightweight modules to perform CSLR. }
   \label{fig2}
\end{figure}

\subsection{Overview}
Given a sign language video with $T$ frames $x = \{x_{t}\}_{t=1}^T \in \mathcal{R}^{T \times 3\times H \times W} $, a CSLR model aims to translate the input video into a series of glosses $y=\{ y_i\}_{i=1}^{N}$ to express a sentence, with $N$ denoting the length of the sentence. The CTC loss~\cite{graves2006connectionist} is used to provide supervision during training by aligning input frames to ground-truth gloss sequences. We follow recent CSLR methods~\cite{Min_2021_ICCV,cheng2020fully, hao2021self,hu2022temporal,hu2023self} to first deploy a spatial extractor to extract frame-wise features, and then employ a sequence model to perform temporal reasoning for sentence prediction. Fig.~\ref{fig2} shows the framework overview of our AdaptSign. While large-scale vision-language models have shown excellent performance over a series of downstream tasks~\cite{dzabraev2021mdmmt, kim2021vilt,radford2021learning,gabeur2020multi,li2022grounded,alayrac2022flamingo,li2022blip,li2023blip}, directly fine-tuning a large-scale model over downstream tasks easily leads to inferior performance and unstable generalizability~\cite{ju2022robust}, which hinder them from broader applications when faced with new data or tasks. Besides, their high computational demands may not always be available in most real-world applications due to constrained computing resources. To successfully transfer their powerful features into tasks with limited available data like CSLR, we propose a novel strategy, termed AdaptSign. Specifically, to keep high training efficiency, we adopt a \textit{frozen} CLIP model~\cite{radford2021learning} as the spatial extractor, and stack several lightweight learnable modules on top to adapt its general features for downstream task target. Specially, we propose attention \& FFN adaption, multi-scale aggregation as well as prefix embedding to inject task-related spatial information into the well-generalized features. To model the temporal movements in sign videos, a cross-frame attention module is introduced to capture the trajectories of the signer. Finally, we add the features of both cross-frame attention and multiscale aggregation as the output representation, and feed them into the sequence model for sentence prediction.  

\subsection{Obtaining More Representative Features}
As a specific language with its own grammar and lexicon~\cite{dreuw2007speech,ong2005automatic} expressed through both manual components (hand/arm gestures) and non-manual components (facial expressions, head movements, and body postures), sign language requires special expert knowledge to understand. We argue the general features of CLIP acquired by pretraining on large-scale web data don't directly generalize well over this specialized domain. Thus, we introduce a series of lightweight modules to adapt the frozen CLIP backbone to learn specific spatial sign features. Besides, as the CLIP model lack the ability to model local temporal information (e.g., cross-frame trajectories of hand/face for the signer), we introduce a cross-frame attention module to encode temporal information into the frozen CLIP backbone.

\subsubsection{Attention \& FFN adaption.}
Efficient transfer learning methods~\cite{houlsby2019parameter,hu2021lora,he2021towards,lester2021power,liu2021p,li2021prefix}  have been extensively explored in Natural Language Processing (NLP) tasks to adapt large language models for downstream applications, which can achieve comparable or even superior performance compared to fine-tuning the whole heavy model. This set of works can be roughly divided into two categories, including Adapters~\cite{houlsby2019parameter,hu2021lora,he2021towards} and Prompt-tuning~\cite{lester2021power,liu2021p,li2021prefix}. Inspired by the achievements of efficient fine-tuning techniques in NLP, we introduce Adapter~\cite{houlsby2019parameter} to learn generalized visual features.

Specifically, the Adapter consists of two fully connected (FC) layers and a GELU activation function in between, with skip connections in parallel. Especially, to lower the required computations and parameters, the first FC layer reduces the channels by a factor of $r$, and the second FC layer project it back to the original dimension. To reuse the learned ability on modelling patch relationships, we add an Adapter in parallel with the LN-MSA layer to adapt the pretrained image features for videos. During training, all other layers in the backbone are frozen and only the Adapters are updated. The outputs of the Adapter and MSA layer are added to calibrate general visual features with specialized video information. The computing procedure could be expressed as follows, with $z_{l-1}$ denoting the features of $(l-1)_{th}$ layer:

\begin{equation}
   \label{e1}
      z_{l}^{'} = z_{l-1} + {\rm MSA}({\rm LN}(z_{l-1})) + {\rm Adapter}(z_{l-1}),
   \end{equation}

To keep the original behavior of the image model, we initialize the weights of the second FC layer in the Adapter with zeros, Thus, the Adapter would act as an identity function at the beginning of training, without hurting the learned features of the pretrained image model. Practically, as the Adapters are quite lightweight, the incurred extra parameters and computations are few (0.1\%) compared to the frozen backbone. However, as we show in the experiments, this simple design greatly facilitates the potential of large-scale pretrained image models in videos, and well works with limited available data.

Some works~\cite{jia2022visual,bahng2022visual} find the feedforward network (FFN) could learn fine-grained image features by sequentially transforming patches via non-linear transformations. To likewise leverage such encouraging ability of pretrained image models, we add an Adapter in parallel with the LN-FFN layer to calibrate its features to learn specialized video information. The calculation procedure could be expressed as:

\begin{equation}
   \label{e2}
      z_{l} = z_{l}^{'} + {\rm FFN}({\rm LN}(z_{l}^{'}))+ {\rm Adapter}(z_{l}^{'}).
   \end{equation}
The LN and FFN layers are kept frozen and only the Adapters are updated. 

\subsubsection{Prefix embedding.}
While the pretrained models contain generic powerful knowledge acquired by training upon a large corpus, there may lack a proper way to inject specific domain knowledge for them to adapt to specialized downstream tasks. To handle this challenge, we try to inject specific sign knowledge into pretrained image models by appending learnable prefix embeddings in the basic blocks to provide instructions for these impressive models.  

Specifically, in MSA, $x$ first undergoes three linear transformations $W_q$, $W_k$ and $W_v$ to obtain the query $Q$, key $K$ and value $V$. To structurally encode specific sign knowledge into pretrained image models, we append learnable prompt embeddings $P\in \mathcal{R}^{m\times d}$ with the length of $m$ before key $K$ and value $V$, respectively, to reformulate them as $K^{'} = [P;K]$ and $V^{'} = [P;V]$. Here, $d$ denotes the number of intermediate channels. Taking $i_{th}$ patch as an example, the attention operation to compute the output $Y_i$ could be formulated as:

\begin{equation}
   \label{e3}
   Y_{i} = {\rm softmax}(\frac{Q_{i} \cdot [P;K]}{\sqrt{d}})\cdot[P;V].
   \end{equation}
It's noticed that $Q_i$ first computes its affinities with $P$ and then aggregates task-specific information from it across various inputs, Irrelevant to input features. $P$ is individually set and learned across different layers, which is expected to offer specialized knowledge of various spatial hierarchies to update the intermediate features of each CLIP block. Practically, $P$ is randomly initialized, and then updated together with the Adapters via backward gradient propagation, keeping other network components frozen. As the length $m$ of $P$ is often quite small (e.g., 8), the incurred extra parameters are few (\textless 0.1\%) compared to the frozen backbone. 

\subsubsection{Multiscale aggregation.}
Features across different layers are shown to contain beneficial information of various spatial hierarchies~\cite{huang2017multi,fan2021multiscale,lin2017feature,he2015spatial}. To effectively leverage this multiscale information for task target, we progressively aggregate features from different CLIP blocks into a robust unified representation. Specifically, we initialize a new token $x_{mul}$ as a query, and treat features from each frozen CLIP block as keys and values to perform MSA. An MLP layer with LN and skip connections is then employed on $x_{mul}$ to update its features. Taking $l_{th}$ layer as an example, this procedure could be expressed as:

\begin{equation}
   \label{e4}
   x_{mul}^{l'} = x_{mul}^{l} + {\rm MSA}({\rm LN}(x_{mul}^{l}, x_l)),
   \end{equation}

\begin{equation}
   \label{e5}
   x_{mul}^{l+1} = x_{mul}^{l'} + {\rm MLP}({\rm LN}(x_{mul}^{l'})).
\end{equation}
This procedure is repeated for \{1, $\cdots$, $L$\}$_{th}$ CLIP blocks.

Especially, the MSA and MLP operation are only conducted on $x_{mul}$, and thus the overall computing complexity is $\mathcal{O}(1)$ with respect to the patch number $n$ in contrast to the $\mathcal{O}(n^2)$ complexity of CLIP blocks. Overall, the extra computations are few (1.9\%) compared to the CLIP backbone.


\subsubsection{Cross-frame attention.}
Hand and face play a major role in expressing sign language by delivering messages through horizontal/vertical hand movements, finger activities, and facial expressions~\cite{dreuw2007speech,ong2005automatic}. However, the pretrained CLIP model fails to capture the temporal features of these body parts. To depict such trajectories, we compute attention maps between patches in a local temporal neighborhood to get their temporal correspondences. Specifically, we use $x_{cls}^{''} \in \mathcal{ R}^{1 \times d}$ as a query for each frame, and treat neighboring spatial-temporal patches $x_{\tau} = \{ x_{-\tau}, \cdots, x_{\tau } \} \in \mathcal{ R}^{(2\tau+1) \times n\times d}$ within $2\tau$+1 adjacent frames as keys and values to compute attention maps $A \in \mathcal{ R}^{1 \times (2\tau+1) \times n}$ between $x_{cls}^{''}$ and $x_{\tau }$ as:
\begin{equation}
   \label{e6}
   A = {\rm LN}(x_{cls}^{''})\times {\rm LN}(x_{\tau})^{T}.
   \end{equation}

$A$ is then passed through a sigmoid function to generate weights within (0,1) to measure the importance of each neighboring patch. We further subtract its values from 0.5 into the range of $[-0.5, 0.5]$, where informative patches are expected to be emphasized with positive values and unnecessary patches are suppressed with negative values. Next, we element-wisely multiply $A$ with $x_{\tau}$ to aggregate motion information from neighboring spatial-temporal patches $x_{\tau}$, whose results are averaged over all patches and added upon $x_{cls}^{''}$ to encoder temporal correspondences via a residual way as:

\begin{equation}
   \label{e7}
   \widehat{x}_{cls} = \frac{1}{(2\tau+1) n} \sum_{i=1}^{ (2\tau+1)}\sum_{j=1}^{ n}(A^{ij}-0.5)\odot x_{\tau}^{ij} + x_{cls}^{''}
   \end{equation}

As the MSA and aggregation operators in eq.~\ref{e6} and eq.~\ref{e7} are only conducted between a single token $x_{cls}^{''}$ and neighboring patches $x_{\tau}$, the overall computing complexity is $\mathcal{O}(1)$ with respect to the patch number $(2\tau+1)\times n$. Practically, the extra computations brought by the cross-frame attention module are few (1.0\%) compared to the CLIP backbone. 

\subsection{Complexity Analysis}
Totally, the overall extra computations raised by our proposed modules are few (3.2\%) with respect to the frozen CLIP backbone. Despite being efficient, as we show below, our AdaptSign shows clear advantages compared to either the frozen or fine-tuned CLIP backbone upon accuracy.

\section{Experiments}
\subsection{Experimental Setup}
\subsubsection{Datasets.} \textbf{PHOENIX14}~\cite{koller2015continuous} and \textbf{PHOENIX14-T}~\cite{camgoz2018neural} are both recorded from German weather forecast broadcasts before a clean background with a resolution of 210 $\times$ 260. They contain 6841/8247 sentences with a vocabulary of 1295/1085 signs, divided into 5672/7096 training samples, /519 development (Dev) samples and 629/642 testing (Test) samples.

\textbf{CSL-Daily}~\cite{zhou2021improving} revolves the daily life, recorded indoor at 30fps by 10 signers. It contains 20654 sentences, divided into 18401 training samples, 1077 development (Dev) samples and 1176 testing (Test) samples. 

\textbf{CSL}~\cite{huang2018video} is collected in the laboratory environment by fifty signers with a vocabulary size of 178 with 100 sentences. It contains 25000 videos, divided into training and testing sets by a ratio of 8:2.

\subsubsection{Training details.} For fair comparisons, we follow the same setting as state-of-the-art methods~\cite{Min_2021_ICCV,hu2023self} to prepare our model and also restrict the training procedure to a single graphical card. We adopt ViT-B/16~\cite{dosovitskiy2020image} as the spatial extractor with pretrained weights from CLIP~\cite{deng2009imagenet}. The sequence model is consisted of a 1D CNN and a two-layer BiLSTM module, followed by a fully connected layer for prediction. The 1D CNN consists of a sequence of \{K5, P2, K5, P2\} layers where $K$ and $P$ denotes a 1D convolutional layer and a pooling layer with kernel size of 5 and 2, respectively. We train our model for 40 epochs with initial learning rate 0.0001 decayed by 5 after 20 and 30 epochs. Adam optimizer is adopted with weight decay 0.001 and batch size 2. All frames are first resized to 256$\times$256 and then randomly cropped to 224$\times$224, with 50\% horizontal flip and $\pm$20\% random temporal scaling during training. During inference, a central 224$\times$224 crop is simply selected. We use VE and VA losses from VAC~\cite{Min_2021_ICCV} for extra supervision. 

\subsubsection{Evaluation Metric.} We use Word Error Rate (WER) as the evaluation metric, which is defined as the minimal summation of the \textbf{sub}stitution, \textbf{ins}ertion, and \textbf{del}etion operations to convert the predicted sentence to the reference sentence, as:
\begin{equation}
\label{e11}
\rm WER = \frac{ \#sub+\#ins+\#del}{\#reference}.
\end{equation}
Note that the \textbf{lower} WER, the \textbf{better} accuracy.

\begin{table}[t]   
   \centering
   \begin{tabular}{cccc}
   \hline
   Configurations & Step time (s) & Dev(\%) & Test(\%)\\
   \hline
   Freezing & 0.28  & 28.6 & 29.7\\
   fine-tuning& 0.65 (2.34$\times$) & 34.3 & 34.9\\
   Partial-1 & 0.31 & 23.4 & 23.1\\
   Partial-2 & 0.34 &  23.1 & 22.8\\
   Ours & 0.31 (1.15$\times$) & \textbf{18.5} & \textbf{19.4} \\
   \hline
   \end{tabular}
   \caption{Results for different training settings of the spatial extractor, measured with a 3090 graphical card.} 
   \label{tab1} 
   \vspace{-5px}
\end{table}

\subsection{Ablation Study}
We perform ablation studies on both development (Dev) and testing (Test) sets of the PHOENIX14 dataset to verify the effectiveness of AdaptSign.

\textbf{Study on different training paradigms.} We compare our method upon both effectiveness and efficiency against freezing or fine-tuning the CLIP backbone in tab.~\ref{tab1}. Here, Partial-$n$ denotes only partially fine-tuning the last $n$ layers of pretrained models. It's observed that a frozen backbone achieves considerable recognition accuracy (28.6\% \& 29.7\% WER on both sets), while directly fine-tuning it leads to unsatisfactory results (-5.7\% \& -5.2\% compared to freezing), which indicates the difficulty to directly adapt large models upon limited training data. Only fine-tuning the last several layers notably improve the recognition performance. Our method achieves much higher accuracy than them (e.g., +10.0\%, +10.3\% than freezing and +15.8\%, +15.5\% than fine-tuning). Regarding training costs, fine-tuning owns much longer (2.34$\times$) step time per batch compared to freezing and partially fine-tuning. Our method only brings 0.15$\times$ additional step time and consumes comparable step time compared to freezing and partially fine-tuning, which shows much better training efficiency than fine-tuning. Overall, our method demonstrate a much better accuracy-computation trade-off than commonly-used fine-tuning methods.   

\textbf{Study on the effectiveness of each component.} We add the proposed modules one by one on top of the frozen CLIP backbone to verify their effectiveness in tab.~\ref{tab2}. By adding the Attention \& FFN Adaption, a significant +6.0\% \& +6.5\% accuracy boost is observed, which shows the necessity to inject specific sign video representations into the intermediate features of the pretrained visual backbone. Sequentially adding the other three modules could also give +1.4\% \& +1.1\%, +1.4\% \& +1.5\%, +1.6\% \& +2.2\% accuracy boost, which demonstrate the effectiveness to inject domain knowledge, leverage multiscale features and capture cross-frame trajectories.

\begin{table}[t]   
   \centering
   \begin{tabular}{ccccc}
   \hline
   Configurations & Dev(\%) & Test(\%)\\
   \hline
    - & 28.6 & 29.7\\
    w/ attention \& FFN adaption & 22.6 & 23.2 \\
    w/ prompt embedding & 21.2 & 22.1 \\
    w/ multiscale aggregation & 19.8 & 20.6 \\
    w/ cross-frame attention & \textbf{18.5} & \textbf{19.4} \\
   \hline
   \end{tabular}
   \caption{Effectiveness of each component by adding them one by one upon a frozen CLIP backbone.} 
   \label{tab2} 
   \vspace{-8px}
   \end{table}

\textbf{Study on the configurations of attention \& FFN adaption.} The upper part of tab.~\ref{tab3} tests the dimension $r$ of intermediate features between the two FC layers ${\rm fc}_1$ and ${\rm fc}_2$ in Adapters. Intuitively, smaller $r$ always leads to fewer computations with higher WER. We try to find an accuracy-computation trade-off for $r$. Practically, it's observed larger $r$ consistently achieves better performance, which reaches a peak after equalling $\frac{1}{4} d$. We thus set $r=\frac{1}{4} d$ by default. We then test the choice of ${\rm fc}_2$ initialization. It's found that compared to the normal distribution, zero-initializing ${\rm fc}_2$ is critical to obtain high-quality visual features, by keeping the original behaviors of the CLIP backbone at the beginning of training. 

\begin{table}[t]   
   \centering
   \setlength\tabcolsep{5pt}
   \begin{tabular}{ccccc}
   \hline
   Configurations & Dev(\%) & Test(\%)\\
   \hline
   $r$ = $\frac{1}{16}$ & 18.9 & 19.9  \\
   \vspace{2px}
   $r$ = $\frac{1}{4}$ & \textbf{18.5} & \textbf{19.4}  \\
   \vspace{2px}
   $r$ = $\frac{1}{2}$ & 18.6 & 19.6  \\
   \vspace{2px}
   $r$ = $1$ & 18.8 & 19.8 \\
   \hline
   Initialize ${\rm fc}_2$ by normal distribution & 20.3 & 20.6  \\
   Initialize ${\rm fc}_2$ by all zeros & \textbf{18.5} & \textbf{19.4}  \\
   \hline
   \end{tabular}
   \caption{Ablations for the configurations of Adapters in Attention \& FFN Adaption.} 
   \label{tab3} 
   \vspace{-5px}
   \end{table}

\textbf{Study on the configuration of prompt embedding.} We first explore the length $m$ of prompt embeddings in the upper part of tab.~\ref{tab4}. It's observed that the accuracy consistently promotes as $m$ increases, which reaches the peak when $m=8$. We thus set $m=8$ by default. We then test whether to share the prompt embeddings $P$ across layers in the bottom part of tab.~\ref{tab4}. It's noticed that setting independent prompt embeddings for each layer achieves better accuracy. This could be attributed from that independent prompt embeddings in different layers offer specific sign knowledge of various hierarchies to help understand sign videos.
\begin{table}[t]   
   \centering
   \begin{tabular}{ccccc}
   \hline
   Configurations & Dev(\%) & Test(\%)\\
   \hline
    $m=2$ & 20.1 & 20.3 \\
    $m=4$ & 19.8 & 20.0 \\
    $m=8$ & \textbf{18.5} & \textbf{19.4}  \\
    $m=12$ & 19.6 & 19.9  \\
   \hline
   Shared across layers & 20.3 & 20.6  \\
   Independent across layers & \textbf{18.5} & \textbf{19.4}  \\
   \hline
   \end{tabular}
   \caption{Ablations for the configuration of Prefix Adaption.} 
   \label{tab4}
   \vspace{-5px} 
   \end{table}

\begin{table}[t]   
   \centering
   \begin{tabular}{ccccc}
   \hline
   Configurations & Dev(\%) & Test(\%)\\
   \hline
    unidirectional & 19.4 & 20.1  \\
    bidirectional & \textbf{18.5} & \textbf{19.4}  \\
   \hline
   $\tau$=0  & 19.8 & 20.6  \\
   $\tau$=1 & 19.1 & 20.3  \\
   $\tau$=2 & \textbf{18.5} & \textbf{19.4}  \\
   $\tau$=3 & 18.9 & 19.9  \\
   \hline
   \end{tabular}
   \caption{Ablations for the configurations of Cross-Frame Attention.} 
   \label{tab5} 
   \vspace{-5px}
   \end{table} 
\begin{figure*}[t]
    \centering
    \includegraphics[width=\linewidth]{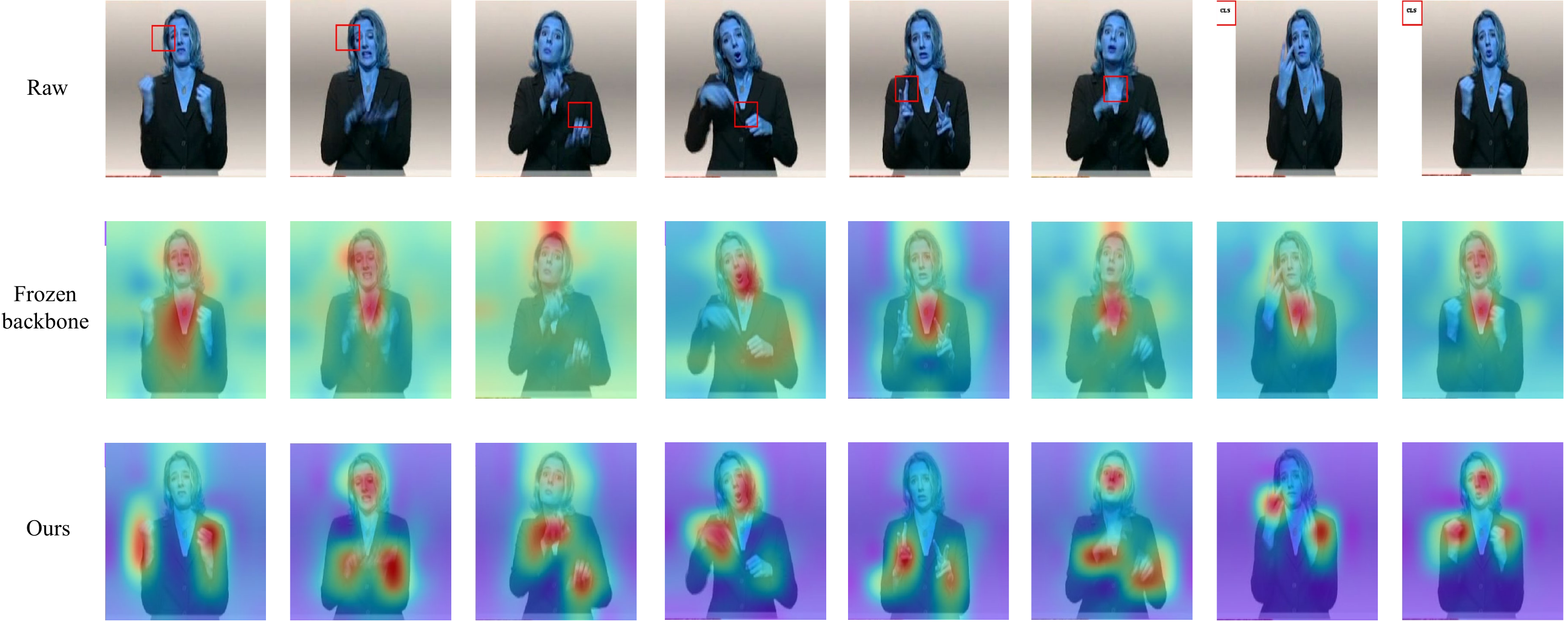}
    \caption{Visualizations of spatial attention maps. Top: raw frames; Middle: attention maps of a frozen CLIP backbone; Bottom: attention maps of ours. The red box denotes the query location. Compared to the frozen CLIP backbone, our method could learn to emphasize domain-sensitive regions like hands and face (dark red areas) which play an important role in expressing sign language.}
    \label{fig4}
    \end{figure*} 
\textbf{Study on the configurations of cross-frame attention.} We first explore whether to perform cross-frame attention bidirectionally or uni-directionally. As shown in the upper part of tab.~\ref{tab5}, aggregating temporal information from bidirectional frames outperforms unidirectional attention observing only future frames, which demonstrates the benefits of both past and future information. We then explore the temporal neighborhood $\tau$ of cross-frame attention. Here, $\tau=0$ means no temporal information is incorporated. In the bottom part of tab.~\ref{tab5}, it's first observed larger $\tau$ consistently brings better performance. When $\tau$ reaches 2, it brings no more performance gain. We set $\tau$ as 2 by default.

\begin{table}[t]   
   \centering
   \begin{tabular}{cccc}
   \hline
   Backbone & Configuration & Dev(\%) & Test(\%)\\
   \hline
  \multirow{3}{*}{CLIP~\cite{radford2021learning}} & Freezing & 29.7 & 30.6\\
   &Partial-1 & 24.2 & 23.8 \\
    & AdaptSign & \textbf{19.5} & \textbf{19.8}  \\
    \hline
    \multirow{3}{*}{CoCa~\cite{YuWVYSW22}} &Freezing & 29.4 & 29.9\\
   &Partial-1 & 23.8 & 23.2 \\
   & AdaptSign & \textbf{19.1} & \textbf{19.4}  \\
   \hline
   \end{tabular}
   \caption{Deploying AdaptSign over multiple backbones on the PHOENIX14 dataset, with ViT-B/32 as the spatial extractor.} 
   \label{tab_flex} 
   \vspace{-7px}
   \end{table}

\textbf{Flexibility over multiple backbones.}
We verify the flexibility of our AdaptSign by deploying it over multiple large-scale vision-language backbones, such as CLIP~\cite{radford2021learning} and CoCa~\cite{YuWVYSW22}, with ViT-B/32 as the spatial extractor. The results are listed in tab.~\ref{tab_flex}. It's observed our AdaptSign generalizes well across different backbones to improve the performance of CSLR.
  
\subsection{Visualizations}   
\textbf{Visualizations for spatial attention maps.} Fig.~\ref{fig4} compares the attention maps generated by the last layer of our method with those from the last layer of the frozen CLIP visual backbone. Notably, our method could generally focus on the human body (light yellow areas), and pays specific attention to regions like hands and face (dark red areas) which play an important role in expressing signs. Instead, the attention maps of the frozen CLIP backbone are much sparser, which mostly focus on static objects like clothes or backgrounds. These results show that our method could help the frozen CLIP model learn to emphasize more specific information in expressing signs, e.g., fine-grained features of hands and face to understand sign language. 

\begin{figure}[t]
  \centering
  \includegraphics[width=\linewidth]{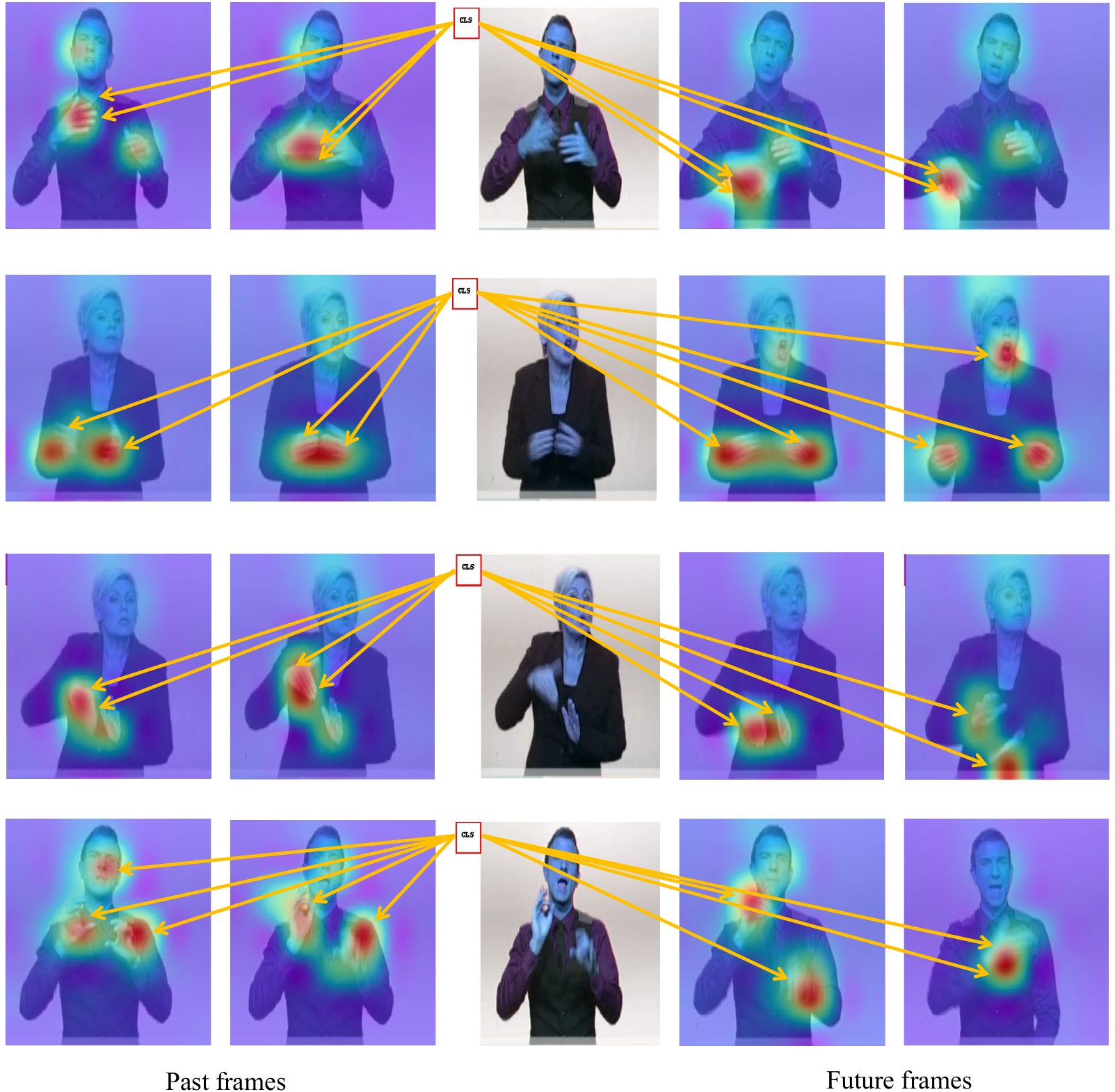}
  \caption{Visualizations of attention maps generated by our cross-frame attention module. The red box denotes the query location, i.e., the frame-level token $x_{cls}^{''}$. It's observed that the query could always attend to informative regions in neighboring frames, e.g., hands or face, to track critical body trajectories in expressing a sign.}
  \label{fig5}
  \vspace{-5px}
  \end{figure}

\textbf{Visualizations for the cross-frame attention module.} Fig.~\ref{fig5} shows the attention maps generated by our cross-frame attention module. The red box denotes the query location, i.e., the frame-level token $x_{cls}^{''}$. It's observed that the query could always attend to informative regions in neighboring frames, e.g., hands or face, to track critical body trajectories in expressing a sign. Especially, it learns to pay special attention to the moving body parts that play a major role in expressing signs. For example, for figures in the first row, the query (left hand) always pays major attention to the quickly moving right hand to capture its trajectories across frames, but pays much less attention to the static hand (left hand itself) and other regions. 

\begin{table}[t]   
   \centering
   \small
   \setlength\tabcolsep{3pt}
   \begin{tabular}{ccccc}
   \hline
   \multirow{2}{*}{Methods} & \multicolumn{2}{c}{PHOENIX14} & \multicolumn{2}{c}{PHOENIX14-T} \\
   &Dev(\%) & Test(\%) & Dev(\%) & Test(\%)\\
   \hline
   FCN~\cite{cheng2020fully} & 23.7 & 23.9 & 23.3& 25.1\\
   CMA~\cite{pu2020boosting} & 21.3 & 21.9  & -&-\\
   VAC~\cite{Min_2021_ICCV} & 21.2 & 22.3 &- &-\\
   SMKD~\cite{hao2021self} &20.8 & 21.0 & 20.8 & 22.4\\
   TLP~\cite{hu2022temporal}  & 19.7 & 20.8 & 19.4  & 21.2 \\
   RadialCTC~\cite{min2022deep} & 19.4 & 20.2 & - & -\\
   SEN~\cite{hu2023self} &  19.5 &  21.0 &  19.3 &  20.7 \\
   CTCA~\cite{guo2023distilling} & 19.5 & 20.1 & 19.3 & 20.3 \\
   CoSign-2s~\cite{jiao2023cosign} & 19.7  &20.1 & 19.5 & 20.1\\
   CVT-SLR~\cite{zheng2023cvt}  & 19.8 & 20.1 & 19.4 & 20.3\\
   \hline
   C+L+H$^*$~\cite{koller2019weakly}  &26.0 & 26.0 & 22.1 & 24.1 \\
   DNF$^*$~\cite{cui2019deep}  &23.1 & 22.9 & - & -\\
   STMC$^*$~\cite{zhou2020spatial} &21.1 & 20.7 & 19.6 & 21.0\\
   C$^2$SLR$^*$~\cite{zuo2022c2slr} & 20.5 & 20.4 & 20.2 & 20.4  \\
   \hline
   \textbf{AdaptSign} &\textbf{18.5} & \textbf{18.8}  & \textbf{18.6} & \textbf{19.8} \\
   \hline   
   \end{tabular}  
   \caption{Comparison with state-of-the-art methods on the PHOENIX14 and PHOENIX14-T datasets. $*$ indicates extra clues such as face or hand features are included by additional networks or pre-extracted heatmaps.} 
   \label{tab6}
 \end{table}

\begin{table}[t]   
   \setlength\tabcolsep{2pt}
   \centering
   \begin{tabular}{cccc}
   \hline
   Methods&  Dev(\%) & Test(\%)\\
   \hline
   LS-HAN~\cite{huang2018video}  & 39.0  & 39.4\\
   TIN-Iterative~\cite{cui2019deep}  & 32.8  & 32.4\\
   Joint-SLRT~\cite{camgoz2020sign}  & 33.1  & 32.0 \\
   FCN~\cite{cheng2020fully} & 33.2  & 32.5 \\
   BN-TIN~\cite{zhou2021improving} & 33.6  & 33.1 \\
   CTCA~\cite{guo2023distilling} & 31.3 & 29.4 \\
   SEN~\cite{hu2023self} & 31.1 & 30.7 \\
   CoSign-2s~\cite{jiao2023cosign} & 28.1 & 27.2\\
   \hline
   \textbf{AdaptSign} & \textbf{26.7} & \textbf{26.3} \\
   \hline
   \end{tabular}  
   \caption{Comparison with state-of-the-art methods on the CSL-Daily dataset~\cite{zhou2021improving}.} 
   \label{tab7}
   \end{table}

\begin{table}[h!]   
  \centering
  \begin{tabular}{cc}
    \hline
    Methods&  WER(\%)\\
    \hline
    LS-HAN~\cite{huang2018video}  & 17.3 \\
    SubUNet~\cite{cihan2017subunets}   & 11.0\\
    SF-Net~\cite{yang2019sf} & 3.8 \\
    FCN~\cite{cheng2020fully}   & 3.0 \\
    STMC~\cite{zhou2020spatial}  & 2.1 \\
    VAC~\cite{Min_2021_ICCV} & 1.6 \\
    C$^2$SLR~\cite{zuo2022c2slr} & 0.9 \\
    \hline
    \textbf{AdaptSign} & \textbf{0.7} \\
    \hline
    \end{tabular}  
    \caption{Comparison with state-of-the-art methods on the CSL dataset~\cite{huang2018video}.} 
    \label{tab8}
    \vspace{-5px}
  \end{table} 
\subsection{Comparison with State-of-the-Art Methods}
\textbf{PHOENIX14} and \textbf{PHOENIX14-T}. Tab.~\ref{tab6} comprehensively compare our method and other state-of-the-art approaches. The entries notated with $*$ indicate these methods utilize additional factors like face or hand features for better accuracy. All previous CSLR methods fine-tune the pretrained image backbone to prepare their recognition model. Compared to these methods that require 100\% tuned parameters, our AdaptSign adapts generic features from frozen pretrained models to learning specific sign representations and outperforms them by a large margin upon both datasets. Especially, AdaptSign outperforms previous CSLR methods equipped with hand and faces acquired by heavy pose-estimation networks or pre-extracted heatmaps (notated with *), without relying on extra expensive annotations to extract hand or face features. 

\textbf{CSL-Daily}. CSL-Daily is designed to cover a wide range of daily contents covering family life, social contact and daily communication with the largest vocabulary size (2k) among commonly-used CSLR datasets. Tab.~\ref{tab7} shows that our method achieves new state-of-the-art accuracy upon this challenging dataset with notable progress, demonstrating its effectiveness to handle real-life scenarios like daily communication.

\textbf{CSL}. CSL is a widely-used CSLR dataset recorded indoors. Tab.~\ref{tab8} shows that our method could achieve extremely superior accuracy (0.7\% WER) upon this well-examined dataset, outperforming existing CSLR methods.

\section{Conclusion}
With impressive performance over a series of downstream tasks, the large scale and scarcity of training data limit the application of large vision-language models to downstream tasks. We propose an efficient training strategy to transfer high-quality visual features into CSLR. Our experiments show that despite being efficient, our strategy outperforms previous methods by a large margin across commonly-used benchmarks. Visualizations show that our proposed method could well attend to informative regions in expressing signs like hands and face as well as their trajectories across frames to capture discriminative sign features.

\section*{Acknowledgements}
This study was supported by National Key Research and Development Program of China (2023YFF0906200) and National Natural Science Foundation of China (Grant Nos. 62072334).
{
    \small
    \bibliographystyle{ieeenat_fullname}
    \bibliography{ref}
}


\end{document}

%% file: preamble.tex
\usepackage[dvipsnames]{xcolor}
